\documentclass[runningheads,a4paper]{llncs}

\usepackage{amssymb,amsfonts}
\usepackage{graphicx}
\usepackage{color}
\usepackage{float}
\usepackage{wrapfig}
\usepackage{amssymb}
\restylefloat{figure}
\usepackage{array}
\newcolumntype{P}[1]{>{\centering\arraybackslash}p{#1}}

\usepackage{url}
\urldef{\mailsa}\path|{lvapeab,fcn}@prhlt.upv.es|
\urldef{\mailsb}\path|{marc.bolanos,petia.ivanova}@ub.edu|

\newcommand{\w}{\mathbf{w}}
\newcommand{\x}{\mathbf{x}}

\newcommand{\z}{\mathbf{z}}
\newcommand{\ve}{\mathbf{v}}
\newcommand{\me}{\mathbf{c}}
\newcommand{\out}{\mathbf{o}}
\newcommand{\inp}{\mathbf{i}}
\newcommand{\fo}{\mathbf{f}}
\newcommand{\bias}{\mathbf{b}}
\newcommand{\prob}{\mathbf{p}}

\newcommand{\g}{\mathbf{g}}
\newcommand{\f}{\mathbf{f}}
\newcommand{\h}{\mathbf{h}}

\newcommand{\A}{\mathbf{A}}

\newcommand{\E}{\mathbf{E}}
\newcommand{\U}{\mathbf{U}}
\newcommand{\V}{\mathbf{V}}
\newcommand{\W}{\mathbf{W}}

\newcommand{\R}{\mathbb{R}} 

\newcommand{\keywords}[1]{\par\addvspace\baselineskip
\noindent\keywordname\enspace\ignorespaces#1}

\usepackage{footnote}
\makesavenoteenv{tabular}

\begin{document}

\mainmatter  

\title{Video Description using Bidirectional Recurrent Neural Networks}

\titlerunning{Video Description using Bidirectional Recurrent Neural Networks}

%
%
\author{\'Alvaro Peris$^1$ \and Marc Bola\~nos$^2$$^\textrm{,}$$^3$\and Petia Radeva$^2$$^\textrm{,}$$^3$\and Francisco Casacuberta$^1$}
\authorrunning{\'A.~Peris, M.~Bola\~nos, P.~Radeva and F.~Casacuberta}

\institute{$^1$ PRHLT Research Center, Universitat Polit\`ecnica de Val\`encia, Valencia (Spain)  \\$^2$ Universitat de Barcelona, Barcelona (Spain) \\$^3$ Computer Vision Center, Bellaterra (Spain) \\
\mailsa\\
\mailsb}

%
%

\toctitle{Video Description using Bidirectional Recurrent Neural Networks}
\tocauthor{\'A.~Peris, M.~Bola\~nos, P.~Radeva and F.~Casacuberta}
\maketitle

\begin{abstract}
Although traditionally used in the machine translation field, the encoder-decoder framework has been recently applied for the generation of video and image descriptions. The combination of Convolutional and Recurrent Neural Networks in these models has proven to outperform the previous state of the art, obtaining more accurate video descriptions.
In this work we propose pushing further this model by introducing two contributions into the encoding stage. First, producing richer image representations by combining object and location information from Convolutional Neural Networks and second, introducing Bidirectional Recurrent Neural Networks for capturing both forward and backward temporal relationships in the input frames.

\keywords{Video Description, Neural Machine Translation, Birectional Recurrent Neural Networks, LSTM, Convolutional Neural Networks.}
\end{abstract}

\section{Introduction}

Automatic generation of image descriptions is a recent trend in Computer Vision that represents an interesting, but difficult task. This has been possible due to the dramatic advances in Convolutional Neural Network (CNN) models that allowed to outperform the state-of-the-art algorithms in many computer vision problems: object recognition, object detection, activity recognition, etc. Generating descriptions of videos represents an even more challenging task that could lead to multiple applications (e.g. video indexing and retrieval, movie description for multimedia applications or for blind people or human-robot interaction). We can imagine the amount of data generated in YouTube (every day people watch hundreds of millions of hours of video and generate billions of views) and how video description would help to categorize them, provide an efficient retrieval mechanism and serve as a summarization tool for blind people. 

However, the problem of video description generation has several properties that make it specially difficult. Besides the significant amount of image information to analyze, videos may have a variable number of images and can be described with sentences of different length. Furthermore, the descriptions of videos use to be high-level summaries that not necessarily are expressed in terms of the objects, actions and scenes observed in the images. There are many open research questions in this field requiring deep video understanding. Some of them are how to efficiently extract important elements from the images (e.g. objects, scenes, actions), to define the local (e.g. fine-grained motion) and global spatio-temporal information, determine the salient content worth to describe, and generate the final video description. All these specific questions need the attention of computer vision, machine translation and natural language understanding communities in order to be solved.


In this work, we propose to enrich the state-of-the-art architecture using bidirectional neural networks for modeling relationships in two temporal directions. Furthermore, we test the inclusion of supplementary features, which help to detect contextual information from the scene where the video takes place.
\section{Related Work}

Although the problem of video captioning recently appeared thanks to the new learning capabilities offered by Deep Learning techniques, the general pipeline adopted in these works resembles the traditional encoder-decoder methodology used in Machine Translation (MT). The main difference is that, in the encoder step, instead of generating a compact representation of the source language sentence, we generate a representation of the images belonging to the video.

MT aims to automatically translate text or speech from a source to a target language. Within the last decades, the prevailing approach is the statistical one~\cite{Koehn10}. The application of connectionist models in the area has drawn much the attention of researchers in the last years. Moreover, a new approach to MT has been recently proposed: the so-called Neural Machine Translation, where the translation process is carried out by a means of a large Recurrent Neural Network (RNN)~\cite{Sutskever14}. These systems rely on the encoder-decoder framework: an encoder RNN produces a compact representation of an input sentence in the source language, and the decoder RNN takes this representation and generates the corresponding target language sentence. Both RNNs usually make use of gated units, such as the popular Long Short-term Memory (LSTM)~\cite{Hochreiter97}, in order to cope with long-term relationships.

The recent reintroduction of Deep Learning in the Computer Vision field through CNNs~\cite{krizhevsky2012imagenet}, has allowed to obtain new and richer image representations compared to the traditional hand-crafted ones. These networks have demonstrated to be a powerful tool to extract feature representations for several kinds of computer vision problems like on objects~\cite{ILSVRC15} or scenes~\cite{zhou2014learning} recognition. Thanks to the CNNs ability to serve as knowledge transfer mechanisms, 
they have also been usually used as feature extractors.


The majority of the works devoted to generate textual descriptions from single images also follow the encoder-decoder architecture. In the encoding stage, they apply a combination of CNN and LSTM for describing the input image. In the decoding stage, an LSTM is in charge of receiving the image information and generating, word by word, a final description of the image~\cite{vinyals2015show}. 
%
%
The problem of video captioning is similar. Seminal works applied methodologies inspired by classical MT~\cite{rohrbach2013translating}. 
Nevertheless, more recent works following the encoder-decoder approach, obtained state-of-the-art performances~\cite{venugopalan2015sequence,yao2015describing}. 

We present a new methodology for natural language video description that makes use of deeper structures and a double-way analysis of the input video. We propose to use as a base architecture the one introduced in~\cite{yao2015describing}. On the top of it, our contributions are twofold. First, we produce richer image representations by combining complementary CNNs for detecting objects and contextual information from the input images. Second, we introduce a Bidirectional LSTM (BLSTM) network in the encoding stage, which has the ability to learn forward and backward long-term relationships on the input sequence. Moreover, we make our code public\footnote{https://github.com/lvapeab/ABiViRNet}.

\section{Methodology}

\begin{figure}[h]
\centering
  \vspace{-1em}
  \includegraphics[width=0.9\textwidth, keepaspectratio]{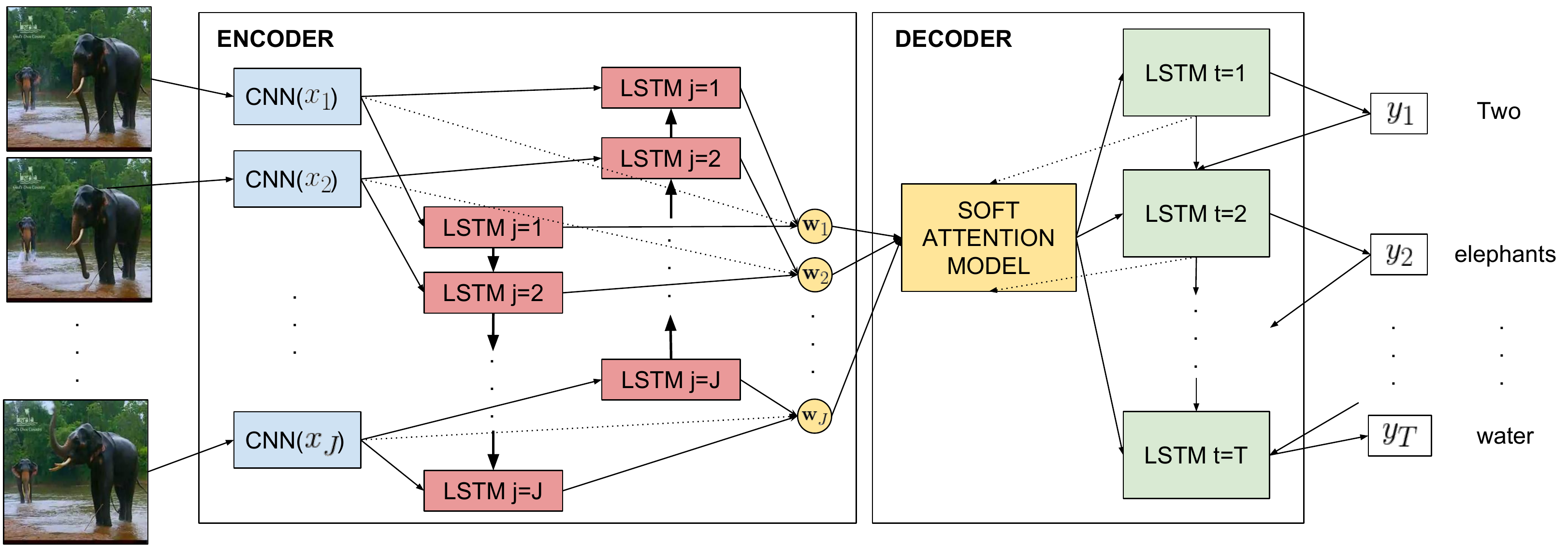}
  \vspace{-1em}
  \caption{\label{fig:methodology_scheme}General scheme of our proposed methodology.}
\end{figure}

An overview of our proposal is depicted in Fig. \ref{fig:methodology_scheme}. We propose an encoder-decoder approach consisting of four stages, using both CNNs and LSTMs for describing images and for modeling their temporal relationship, respectively.

\vspace{-1em}

\textit{First} (blue in the scheme), we apply two state of the art CNN models for extracting complementary features on each of the raw images from the video.

\textit{Second} (red in the scheme), considering we need to describe the actions performed in consecutive frames, we apply a BLSTM for capturing temporal relationships and complementary information by taking a look at the action in a forward and in a backward manner. 

\textit{Third} (yellow in the scheme), the two output vectors from forward and backward LSTM models of the previous step are concatenated together with the CNN output for each image and are fed to a soft attention model in the decoder. This model decides on which parts of the input video should focus for emitting the next word, considering the description generated so far. 

\textit{Fourth} (green in the scheme), an LSTM network generates the video caption from the representation obtained in previous stages. The variable-length caption is obtained word by word, using a softmax function on the top of this LSTM network.

\subsection{Encoder}
\begin{wrapfigure}{R}{0.5\textwidth}
  \vspace{-0.6cm}
\centering
  \includegraphics[width=0.5\textwidth, keepaspectratio]{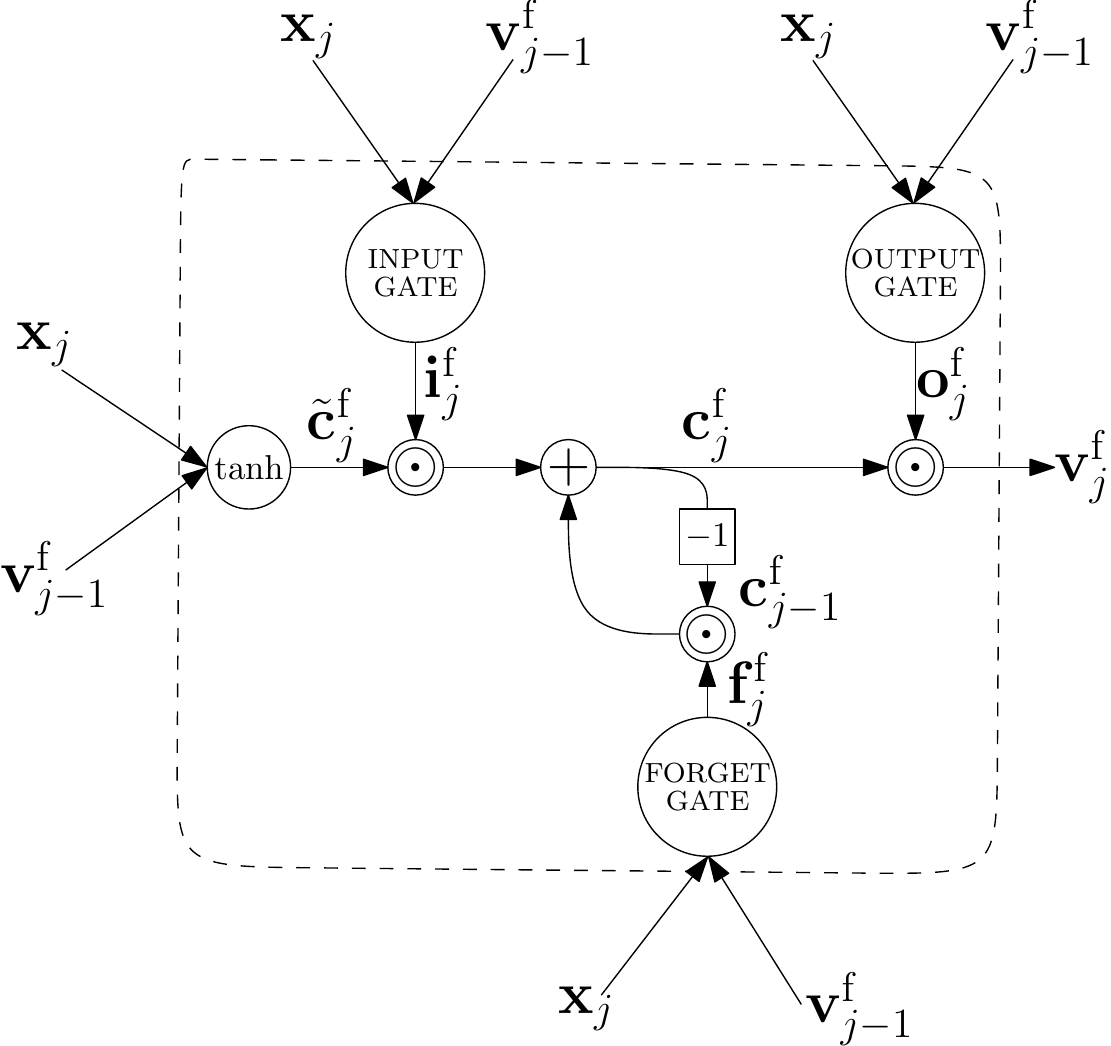}
  \caption{\label{fig:lstme}Forward layer LSTM unit for the encoder. The output depends on the previous hidden state ($\ve^{\mathrm f}_{j-1}$) and the current feature vector from the video extracted by the CNN ($\x_{j}$). Input, output and forget gates module the amount of information that flows across the unit.  \vspace{-0.5cm}}
  \vspace{-0.5cm}
\end{wrapfigure}
  
Given the video description problem, in the encoding stage we need to properly characterize the video for 1) understanding which kind of objects and structures appear in the images, and 2) modeling their relationships and actions along time.

For tackling the first part of the problem, several kinds of pretrained CNNs may be used for describing the images, which can be distinguished by the different architectures or by the different datasets used for training. Although an extended comparison and combinations of models could be used for applying this characterization, we propose combining object and context-related information. For this purpose we use a GoogleNet architecture \cite{szegedy2015going} separately trained on two datasets, one for objects (ILSVRC dataset \cite{ILSVRC15}), and the other for scenes (Places 205 \cite{zhou2014learning}). The combination of these two kinds of data can inform about the objects appearing and their surroundings, being ideal for the problem at hand. For a given video, the CNNs generate a sequence $\V_c$ of $J$ $d$-dimensional feature vectors, $\x_1 , \dots , \x_J$ with $\x_j\in\R^d$ for $1\leq j \leq J$, where $J$ is the number of frames in the video.

To solve the second problem, a BLSTM processes the sequence $\V_c$, generating a new sequence $ \V_{bi} = \ve_1 , \dots , \ve_J$ of $J$ vectors. BLSTM networks are composed of two independent LSTM layers namely, forward and backward. Both layers are analogue, but the latter processes the input sequence reversed in time.

LSTM networks have, in addition to the classical hidden state, a memory state. Let $\ve^{\mathrm f}_j$ be the forward layer hidden state at the time-step $j$, and let $\me^{\mathrm f}_j$ be its memory state. The hidden state $\ve^{\mathrm f}_j$ is computed as $\me^{\mathrm f}_j$ controlled by an output gate ${\out}^{\mathrm f}_j$. The current memory state depends on an updated memory state, and on the previous memory state, $\me^{\mathrm f}_{j-1}$, respectively modulated by the forget and input gates, $\fo^{\mathrm f}_j$ and $\inp^{\mathrm f}_j$. The updated memory state $\tilde{\me}^{\mathrm f}_j$ is obtained by applying a logistic non-linear function to the input and the previous hidden state. Each LSTM gate has associated two weight matrices, accounting for the input and the previous hidden state. Such matrices must be estimated on a training set. Figure~\ref{fig:lstme} shows an illustration of an LSTM unit. The same architecture applies to the backward layer, but dependencies flow from the next time-step to the previous one. Since forward and backward layers are independent, they have different weight matrices to estimate.

Each feature vector $\ve_j$ computed by the BLSTM results as the concatenation of the forward and backward hidden states: $\ve_j = [ \ve^{\mathrm f}_j ; \ve^{\mathrm b}_j ]\in\R^{2\cdot D}$ for $1\leq j \leq J$, being $D$ the size of each forward and backward hidden state. 

Finally, the encoder combines the sequences $\V_{c}$ and $\V_{bi}$ by concatenating the vectors from the CNN and from the BLSTM, producing a final sequence $\V$ of $J$ feature vectors $\w_1 , \dots , \w_J$, $\w_j = [\x_j;\ve_j]\in\R^{d+2 \cdot D}$ for $1\leq j \leq J$.

\subsection{Decoder}
\begin{wrapfigure}{R}{0.5\textwidth}
  \vspace{-3mm}
\centering
  \includegraphics[width=0.5\textwidth, keepaspectratio]{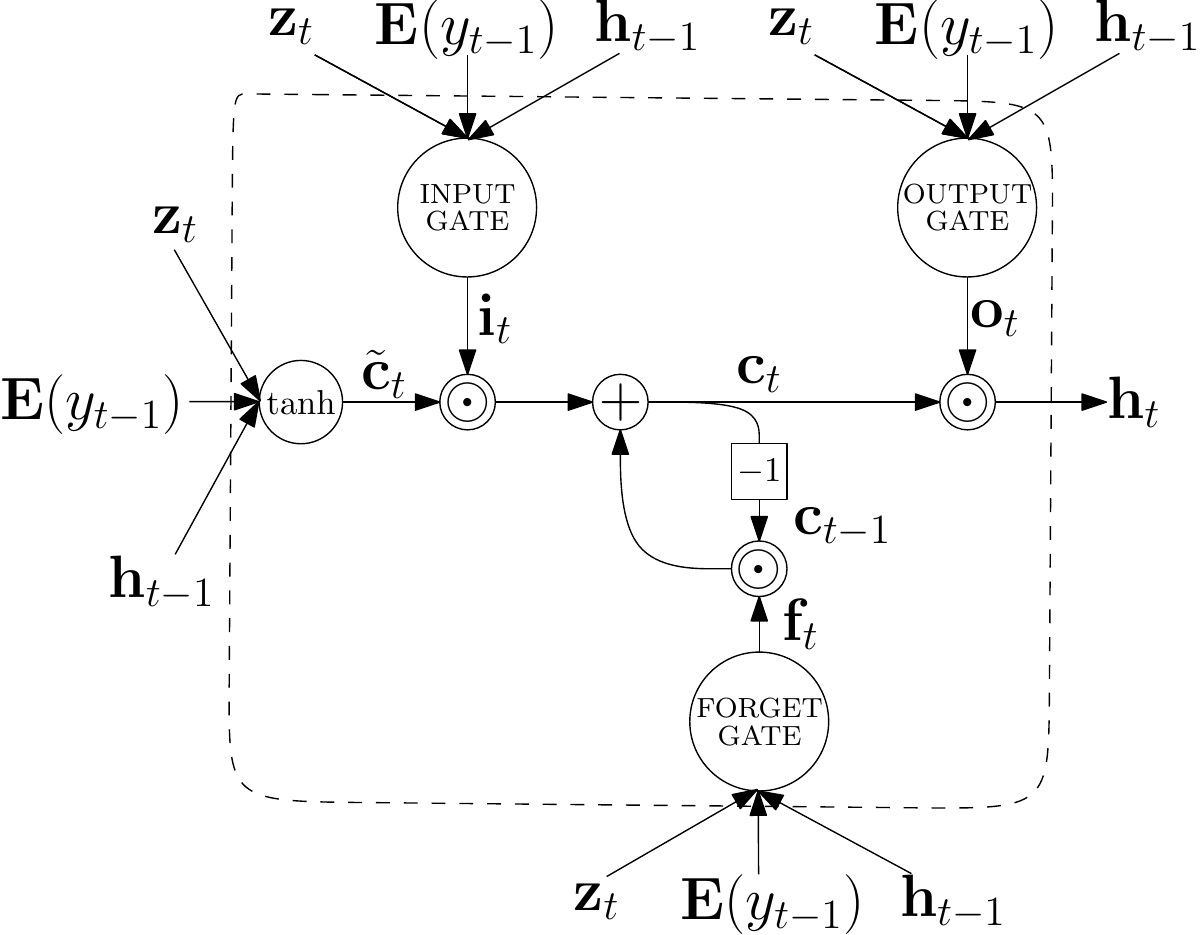}
  \caption{\label{fig:lstm}Decoder LSTM unit. The output depends on the previous hidden state ($\h_t$), the word embedding of the previously generated word ($\E(y_{t-1})$) and the context vector provided by the attention mechanism ($\z_t$).\vspace{-0.5cm}}
  \vspace{-5mm}
\end{wrapfigure}  

The decoder is an LSTM network, which acts as a language model, conditioned by the information provided by the encoder. 
This network is equipped with an attention mechanism~\cite{Bahdanau15,yao2015describing}: a soft alignment model, implemented as a single-layered perceptron, that helps the decoder to know \emph{where} to look at for generating each output word. Given the sequence $\V$ generated by the encoder, at each decoding time-step $t$ the attention mechanism weights the $J$ feature vectors and combines them into a single context vector $\z_t\in\R^{d+2 \cdot D}$.
%
%

The decoder LSTM is defined similarly to the forward layer from the encoder, but it takes into account the previously generated word and the context vector from the attention mechanism, in addition to its previous hidden state. The last word representation is provided by a word embedding matrix $\E \in \R^{m \times V}$, being $m$ the size of the word embedding and $V$ the size of the vocabulary. $\E$ is estimated together with the rest of the model parameters. 


A probability distribution over the vocabulary of output words is defined from the hidden state $\h_t$, by means of a softmax function. This function represents the conditional probability of a word given an input video $\V$ and its history (the previously generated words): $p(y_t|y_1, \dots, y_{t-1}, \V)$. Following~\cite{Sutskever14}, a beam-search method is used to find the caption with highest conditional probability.

\section{Results}

In this section we describe the datasets and metrics used for evaluating and comparing our model to the video captioning state of the art.

\subsection{Dataset}

The \textbf{Microsoft Research Video Description Corpus} (MSVD) \cite{chen2011collecting} is a dataset composed of 1970 open domain clips collected from YouTube and annotated using a crowd sourcing platform. Each video has a variable number of captions, written by different users. We used the splits made by~\cite{venugopalan2015sequence,yao2015describing}, separating the dataset in 1200 videos for training, 100 for validation and the remaining 670 for testing. During training, the clips and each of their captions were treated separately, accounting for a total of more than $80,000$ training samples.

\subsection{Evaluation Metrics}

In order to evaluate and compare the results of the different models we used the standardized COCO-Caption evaluation package \cite{chen2015microsoft}, which provides several metrics for text description comparison. We used three main metrics, all of them presented from 0 (minimum quality) to 100 (maximum quality):

\textbf{BLEU} \cite{papineni2002bleu}: this metric compares the ratio of n-gram structures that are shared between the system hypotheses and the reference sentences.
%


\textbf{METEOR} \cite{lavie2009meteor}: it was introduced to solve the lack of the recall component when computing BLEU. It computes the F1 score of precision and recall between hypotheses and references. 
%
%

\textbf{CIDEr} \cite{vedantam2015cider}: similarly to BLEU, it computes the number of matching n-grams, but penalizes any n-gram frequently found in the whole training set. 

%
%


\subsection{Experimental Results}\label{sec:exp_results}

On all the tests we used a batch size of 64, the learning rate was automatically set by the Adadelta~\cite{Zeiler12} method and, as the authors in \cite{yao2015describing} reported, we applied a frame subsampling, picking only one image every 26 frames for reducing the computational load. The parameters of the network were randomly initialized. An evaluation on the validation set was performed every 1000 updates. The learning process was stopped when the reported error increased after 5 evaluations.

For each configuration we run 10 experiments. At each of them, we
randomly set the value of the critical model hyperparameters. Such hyperparameters and their tested ranges are $m \in [300, 700]$, $| \h_t | \in [1000, 3000]$. When using the BLSTM encoder, we performed an additional selection on $| \ve_j | \in [100, 2100]$.

\begin{table}[th]
  \centering
	\begin{tabular}{ l | P{2cm} P{2cm} P{2cm} }
 Model & BLEU~[\%] & METEOR~[\%] & CIDEr~[\%] \\ \hline \hline
Objects* & 51.5 & 32.5 & 66.0 \\ \hline
Objects + BLSTM          & \textbf{53.6} & \textbf{32.6} & 66.4 \\
Objects + Scenes         & 52.6 & 32.5 & 67.0 \\
Objects + Scenes + BLSTM & 52.8 & 31.3 & \textbf{67.2} 
	\end{tabular}
	\caption{Text generation results for each model on the MSVD dataset. The results below the horizontal line are our proposals. *Model from~\cite{yao2015describing} only with \textit{Object} features evaluated on our system. \label{tab:results}}
\vspace{-2em}
\end{table}

For each configuration, the best model with respect to the BLEU measure on the validation set was selected. In Table \ref{tab:results} we report the results of the best models on the test set. The first row correspond to the result obtained with our system with the object features from ~\cite{yao2015describing}. The configurations reported below the horizontal line are our proposals, where \textit{Scenes} indicates we use scene-related features concatenated to \textit{Objects} and \textit{BLSTM} denotes the use of the additional BLSTM encoder.


\section{Discussion and Conclusions}

Analyzing the obtained results, a clear improvement trend can be derived when applying the BLSTM as a temporal inference mechanism. The BLSTM addition when using \textit{Objects} features allows to improve the result on all metrics, obtaining a benefit of more than 2 BLEU points. Adding scenes-related features also slightly improves the result, although it is not as remarkable as the BLSTM improvement. The combination of \textit{Objects}+\textit{Scenes}+BLSTM offers the best CIDEr performance, nevertheless, this result is slightly below the \textit{Objects}+BLSTM one on the other metrics. This behaviour is probably due to the significant increase on the number of parameters to learn. It should be investigated whether the reduction of the number of parameters by reducing the size of the CNN features, or the use of larger datasets could lead to further improvements.

In conclusion, we have presented a new methodology for natural language video description that takes profit from a bidirectional analysis of the input sequence. This architecture has the ability to learn forward and backward long-term relationships on the input images. Additionally, the use of complementary object and scene-related image features has proved to obtain a richer video representation. The improvements have allowed the method to outperform the state-of-the-art results in the problem at hand.

These results suggest that deep structures help to transfer the knowledge from the input sequence of frames to the output natural language caption. Hence, the next step to take must delve into the application deeper modeling structures, such as 3D CNNs and multi-layered LSTM networks. 

\subsubsection*{Acknowledgments.} This work was partially founded by TIN2012-38187-C03-01, SGR 1219, PrometeoII/2014/030 and by a travel grant by the MIPRCV network. P. Radeva is partially supported by an ICREA Academia’2014 grant. We acknowledge NVIDIA for the donation of a GPU used in this work.

\bibliographystyle{plain}
\bibliography{0_main}
\end{document}